# Deep learning-based automated image segmentation for concrete petrographic analysis

Yu Song [1*], Zilong Huang [2], Chuanyue Shen [1], Humphrey Shi [3*], and David A Lange [4]

[1] Affiliation 1:
Research Assistant, Dept. of Civil and Environmental Engineering, University of Illinois Urbana-Champaign
2145 Newmark Civil Engineering Bldg
205 N. Mathews
Urbana IL 61801

[2] Affiliation 2:
Visiting Scholar, IFP Group, Beckman Institute, University of Illinois Urbana-Champaign
405 N Mathews Ave
Urbana IL 61801

[3] Affiliation 3:
Assistant Professor, Dept. of Computer and Information Science, University of Oregon
258 Deschutes Hall
1477 E. 13th Ave.
Eugene OR 97403

[4] Affiliation 4:
Professor, Dept. of Civil and Environmental Engineering, University of Illinois Urbana-Champaign
2129b Newmark Civil Engineering Bldg
205 N. Mathews
Urbana IL 61801

\* Correspondence: hshi3@uoregon.edu; yusong@ucla.edu

**Abstract:**

The standard petrography test method for measuring air voids in concrete (ASTM C457) requires a meticulous and long examination of sample phase composition under a stereomicroscope. The high expertise and specialized equipment discourage this test for routine concrete quality control. Though the task can be alleviated with the aid of color-based image segmentation, additional surface color treatment is required. Recently, deep learning algorithms using convolutional neural networks (CNN) have achieved unprecedented segmentation performance on image testing benchmarks. In this study, we investigated the feasibility of using CNN to conduct concrete segmentation without the use of color treatment. The CNN demonstrated a strong potential to process a wide range of concretes, including those not involved in model training. The experimental results showed that CNN outperforms the color-based segmentation by a considerable margin, and has comparable accuracy to human experts. Furthermore, the segmentation time is reduced to mere seconds.

## 1. Introduction

Concrete is a complex composite material that plays an essential role in modern construction. During production, the material proportioning and mixing protocol affect its structural, serviceability, and durability performance. As such, concrete quality is largely contingent on the property and distribution of the different phase compositions, mainly aggregates, cement paste, and air voids. Petrographic analysis is a common approach for evaluating concrete quality and predicting its long-term performance. While petrographic analysis is a collective term for a series of tests, it generally involves the examination of different phases on a polished concrete section via various imaging methods like optical microscopy [1], flatbed scanning [2,3], scanning electron microscope (SEM) [4], and energy dispersive X-ray analysis (SEM- EDX) [5]. This kind of analyses has been broadly applied on concrete for crack characterization under various damaging mechanisms such as fire exposure [6,7] and alkali-silica reaction (ASR) [8–10], air void analysis for evaluating freeze-thaw performance [2,11], and phase quantification for aggregate [12–14], paste [15,16] and other components [5,17–19].

In practice, manually conducted petrographic inspections often require massive time involvement, as well as high labor costs. A well-known example is the hardened concrete air void analysis, as specified by the ASTM C457 in the US [11] or EN 480-11 in Europe [20]. Although the testing standard is well established, the tedious testing procedures potentially discouraged the acceptance of this test by practitioners. Furthermore, the repeated visual judgment required for the operator over hours during the inspection causes certain concerns about operator subjectivity [21–23].

Pertinent studies in recent decades have attempted to reduce the manual involvement by fulfilling a color-based image segmentation with image analysis techniques. As a common approach, the RapidAir 457 testing instrument [24] and several other studies [25–27] applied black ink and white powder to create a binary surface to highlight air voids. Subsequent advancements differentiated the paste and aggregate using phenolphthalein dye [2,3,28]. In either case, the scan of the sample surface can be segmented based on the created color contrast, making an automatic C457 measurement possible [3] or other advanced analysis [29,30] possible using simple program script.

However, the color-aided petrographic analysis still has a number of limitations due to the following reasons. First, the surface treatment costs extra time and requires high workmanship. Second, due to the subtle variation of color treatment on different samples, the color threshold needs to be readjusted each time. Third, improper color dyeing can reduce segmentation accuracy. These limitations have long been recognized in concrete studies, and some efforts are seen to improve the segmentation accuracy using additional graphic features. In a study by Werner and Lange, for example, a convolution kernel algorithm extracting the texture contrast was used to differentiate the aggregate and cement paste in SEM images [31].

Recently, scientific research is rapidly reshaped by artificial intelligence. Recent developments in the fields of computer vision and machine learning has achieved significant breakthroughs regarding image segmentation, in which the semantic segmentation (i.e., pixel-leveled classification for an image) is particularly relevant to the objective of petrographic analysis. As a machine learning subset, the rise of deep learning-based semantic segmentation has substantially improved the precision and processing speed of machine visual understanding, attracting attention from various fields, like autonomous driving [32], satellite sensing [33], and medical imaging [34]. Garcia-Garcia et al. reviewed the major deep learning techniques for semantic segmentation [35], among which convolutional neural networks (CNN) is well-recognized for visual imagery with its dominant superiority in accuracy and efficiency. As compared with the color-based methods, this approach implements more sophisticated computational strategies for determining different objects in an image autonomously.

Despite the fruitful outcomes of infusing deep learning in many fields, its potential for concrete petrographic analysis has not been well validated. Aiming at advancing the use of deep learning techniques in concrete research, we investigate the efficacy of using CNN for the concrete image segmentation task. With the practical need of an expeditious petrographic analysis in mind, the ultimate objectives of this study are set out to (i) propose a practical guideline of applying CNN in petrographic analysis and (ii) assess its performance for segmenting concrete samples without the color treatment. As for the implementation, the CNN model was developed based on a well-established algorithm, ResNet-101 [36], and trained with a group of concrete image-label pairs. The segmentation performance was evaluated based on both the training images and a set of new testing images. The CNN segmentations were statistically compared against human recognition, as well as the color-based segmentation. Furthermore, the sensitivity of the ASTM C457 air void parameters to the different segmentation methods was investigated.

In this paper, the Background section first details technical issues about the dyeing treatment for color-based segmentation as observed in previous research. Then, the basics of CNN and its state-of-art for semantic segmentation are briefly reviewed. The Methodology section covers the experimental procedure of sample preparation, details of the CNN model selection, training, testing, and also the accuracy assessment. The test results and important findings regarding the segmentation performance of CNN are then addressed.

## 2. Background

*2.1 Technical limitations of the color method*

Taking advantage of the color reaction of phenolphthalein with the high alkalinity in cement paste (pH > 9), the aggregate and paste phases can be differentiated to facilitate the petrographic analysis [3]. This method, however, cannot be used for carbonated concrete that has a lower pH value, which is common for samples cored from old construction. Even for new concrete samples, it is difficult to achieve the desired color contrast, as carbonation initiates once the fresh sample section is exposed to the environment [2,3]. In practice, it is common to encounter problems when using color to segment aggregate and paste phases, as exemplified in Fig. 1. In Fig. 1a, the periphery of coarse aggregate is contaminated by the pink color; this issue is more observed on light-color aggregates. Fig. 1b shows the uneven coloration of cement paste. Under a higher magnification as shown in Fig. 1c, some fine aggregates are colored to pink, which is related to the semi-transparent nature of sand grains (primarily quartz) [30,31]. Based on our observation, the color dye may infiltrate into the substrate through the interfacial transitional zone (ITZ) around sands, leaving the pinkish visual effect. As such, the color treatment could actually constitute a hindrance to the analysis in certain samples. Despite the obvious challenges, the above problems can be fundamentally avoided if the segmentation can be done directly on the uncolored samples.

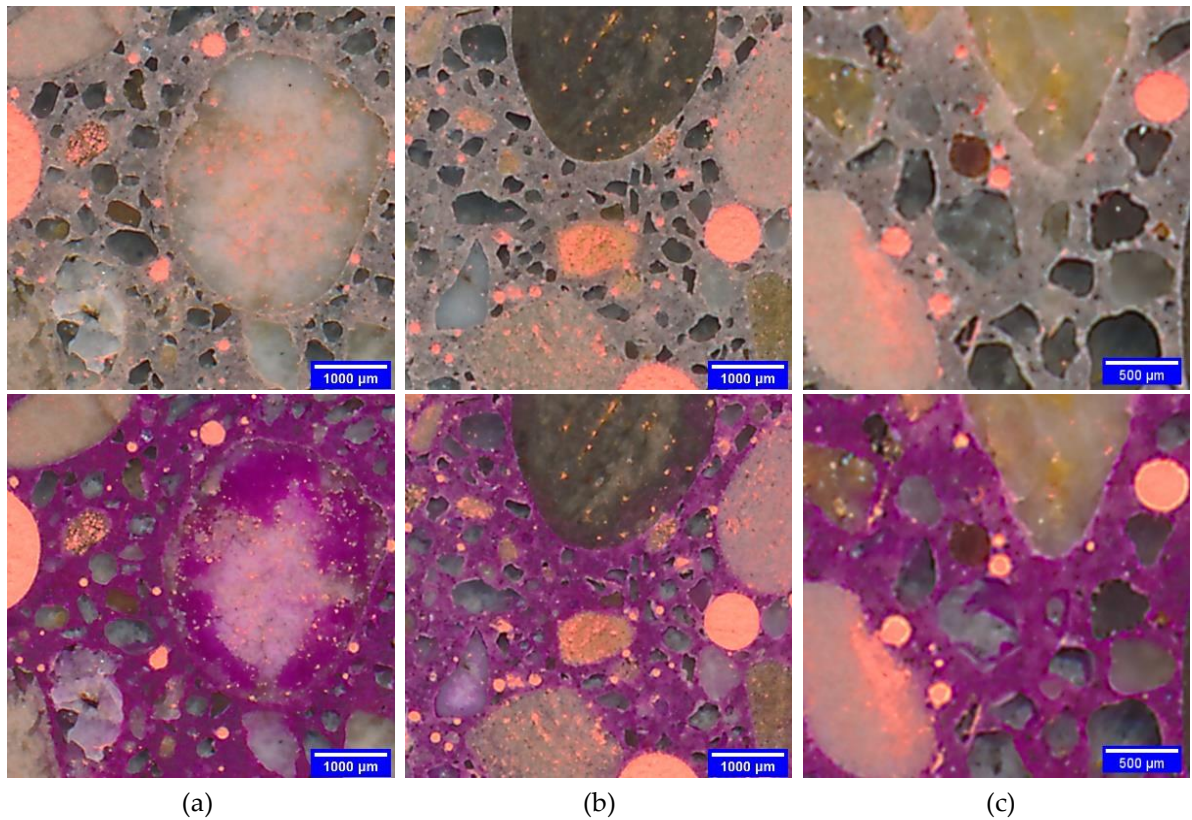

Fig. 1: Typical issues of improper phenolphthalein treatment: (a) color contamination of the coarse aggregate; (b) inhomogeneous color dye of cement paste; (c) color contamination of fine aggregate.

*2.2 Accurate visual understanding using CNN*

Due to the huge success of deep learning models in visual recognition applications, deep learning-based semantic segmentation has emerged as a mainstream research topic in the field of computer vision in recent years. Unlike the color-based segmentation, this new approach relies on a substantial number of high-level features to isolate different regions in an image. Obtained from the nonlinear combinations of low-level features (color, shape, pattern, texture, etc.) that are easy to detect from an image, the high-level features can serve as efficient object descriptors in the program, though they themselves are usually abstract and object-specific, such as a mathematic matrix depicting the facial characteristics of dogs. The state-of-art deep learning technique to fulfill this functionality is CNN, as it can find more discriminative features with less computation [35,37].

CNN is essentially an artificial neural network infused with the concept of convolution kernel—a type of image filtering algorithm commonly used in image processing. An illustration of using CNN for semantic segmentation is given in Fig. 2. In terms of the general structure of CNN, its first layer contains a large number of sublayers with convolutional kernels, which extract hundreds of low-level feature maps from the input image. As the network goes deeper, the subsequent layers in CNN gradually synthesize the low-level features to perceive complex high-level features for a more discriminative decision function. The implementation of CNN for image segmentation can be generally divided into model training and testing. The CNN training typically requires feeding the model with a group of images with their labels. With the iterations, the algorithm updates the model parameters for proposing an optimized segmentation strategy. The above step can be understood as a "learning" process. Afterward, the CNN model can be tested with new images, where a better model is expected to yield more accurate segmentation in this "predicting" process.

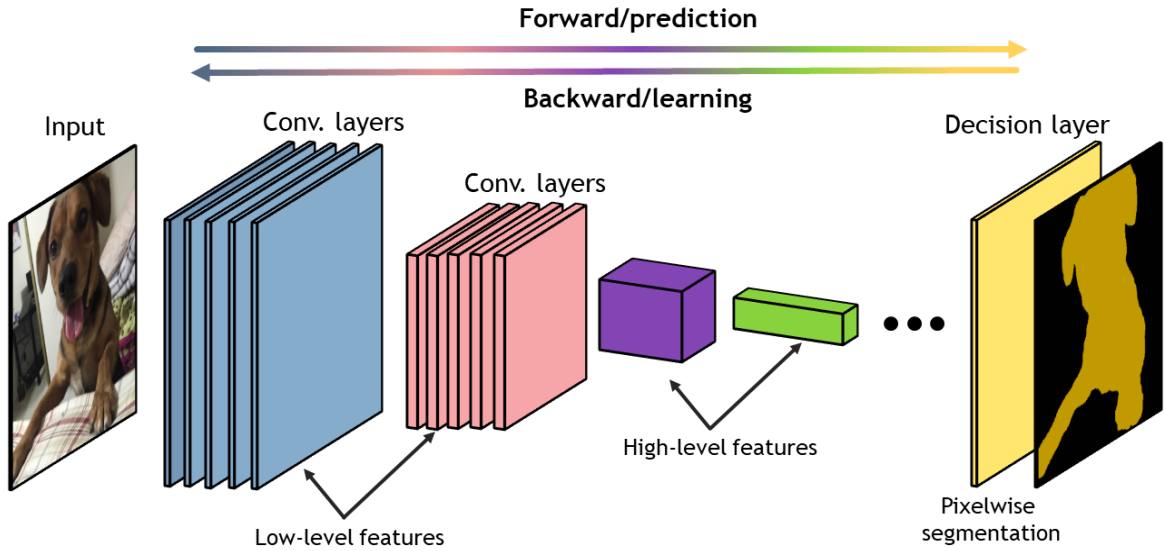

Fig. 2: A simplified CNN structure for semantic segmentation, with an example of dog identification. [Note, this is a conceptual illustration not reflecting the technical details.]

*2.3 Review of the major advancements in deep learning*

Back in 2012, an important work was published by Krizhevsky et al. [38], where the power of deep neural networks was demonstrated with its unprecedented performance in ImageNet Large Scale Visual Recognition Challenge (ILSVRC) competition. In 2014, Simonyan and Zisserman published a fundamental work investigating the effect of the depth of CNN on its accuracy in large-scale image recognition [39]. After comparing different strategies of building CNN, they found that the most effective approach is increasing the CNN layer depth (by that time, up to 19) while using very small convolutional kernels (3×3) in the convolutional layer. This structure configuration, named as VGG by the authors, simultaneously reduces the model parameters involved in the computation and increases the non-linearity of the high-level features.

One of the earliest breakthroughs of using CNN for semantic segmentation was achieved by R. Girshick et al. [37]. The authors proposed a simple and scalable detection algorithm R-CNN, where R stands for region proposals. Specifically, R-CNN first generates category-independent region proposals to highlight locations of interest in the image; then, the feature of each region is computed using a bottom-up CNN structure; and lastly, the features are examined by a classification algorithm such that the regions belonging to different objects can be segmented. In 2014, R-CNN achieved a mean average performance of 53.3% on the canonical VOC evaluation dataset for classification evaluation--a 30% improvement than the previous best. To overcome the need for a large amount of labeled training data in CNN, Pinheiro and Collobert proposed a CNN model that accepts weakly labeled training data (image-level annotation) in 2015 [40]. To correlate with the image-level annotation, the output of this algorithm is an image-level score evaluated based on the pixel-level scores, by using an aggregation algorithm in the last step of the model computation. Although the proposed CNN model is not as accurate as those using fully labeled data, this paper provides innovative insights for reducing the computation cost and bridging semantic segmentation with image classification problems.

Built over the successes of classification neural networks, a fully convolution network (FCN) was proposed by Long et al. in 2015 [41]. This structure first realized an end-to-end and pixel-to-pixel semantic segmentation, where the end-to-end means that the intermediate procedures are not involved with human interference, such as parameter tuning, and pixel-to-pixel means the segmented image can be directly output from the model. To realize these functions, the authors used fully convolutional layers to replace the fully connected layers in several state-of-art CNNs including VGG originally designed for classification problems. The core thoughts of this algorithm are 1) it takes advantage of the powerful feature extractor from the existing CNNs, and then 2) the fully connected layers can, as a reversed manner, deconvolute the high-level features back to an image segmented at pixel-level. It turns out that this idea is extremely successful. The FCN achieved a mIoU (mean of intersection over union, an accuracy measure) of 62.7%, which is a 20% relative improvement than other CNNs with reduced computational demand.

Since CNNs was originally designed for object classification, the emphasis on the invariance of spatial transformations inherently limited the spatial precision for semantic segmentation. To overcome this issue, Chen et al. introduced the idea of conditional random field (CRF), a probabilistic graphical model, into FCN and named this new approach as DeepLab [42]. With the additional implementation of the hole algorithm for sparse feature extraction [43], the author achieved highly refined object boundaries in the segmented image and reduced computational cost. In the continued development of DeepLab, Chen et al. further proposed average-pooling and max-pooling mechanisms to stimulate CNN focusing on more discriminative features [44]. With those improvements, the mIoU scored by DeepLab reached above 70%.

Whereas the greater CNN depth is expected to bring great benefits to segmentation work, the implementation is bottlenecked by a problem known as vanishing gradients. As the CNN layer number increases, the model training becomes a daunting task because the algorithm will eventually stop from convergence, i.e. "learning nothing". With the hypothesis that introducing skip connections to the successive CNN layers can alleviate this problem, He et al. reformulated the classic

CNN layers with a residual learning framework and named it ResNet [36]. In short, this solution ensures that a CNN layer not only receives inputs from its previous layer but also even former layers, such that it should always "learn something". Using this philosophy, the authors successfully modified a VGG structure to reach a maximum layer depth of 152 (ResNet-152), 8 times deeper than the original. With respect to the performance, ResNet beat all previous CNNs and won 1st place in ILSVRC competition in 2015 with 96.4% accuracy, as well as a series of other major machine vision competitions in recent years. Taking advantage of the latest deep learning techniques, ResNet was adopted as the backbone of the CNN model investigated in this study, as detailed in the next section.

## 3. Methodology

The work done in this study can be divided into four parts, as illustrated in Fig. 3. Section 3.1 details the first part, which involves obtaining a group of high-resolution scans from different concrete samples, as well as the corresponding label images via the color-based analysis. Section 3.2 provides the technical information involved with CNN model selection and model training. Section 3.3 covers the CNN testing (i.e., segmenting new concrete scans). Section 3.4 details the strategy used for accuracy assessment.

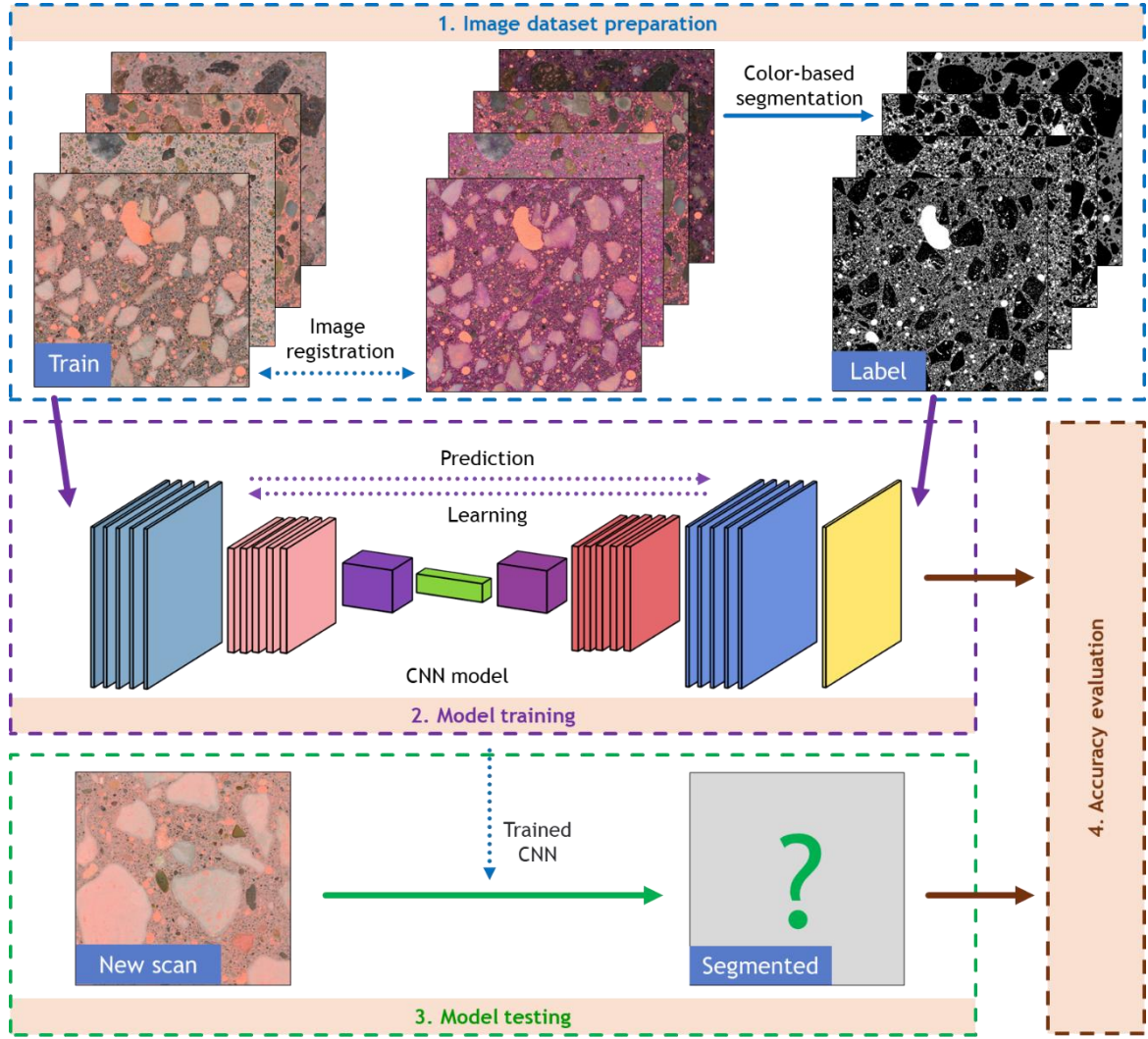

Fig. 3: The framework of the major works conducted in this study. ResNet-101 is adopted as the backbone algorithm for the CNN model.

*3.1 Material and sample preparation*

Eight types of concrete mixtures with various proportioning and material constituents were involved in this study, as summarized in Table 1. The first four mixtures each had two images for training and one for testing, and the other four each had only one for testing. All the samples investigated in this study were obtained in the form of hardened concrete. The labels of the training images were prepared using the color-based segmentation (see Fig. 3). The testing images can be divided into two groups as *familiar* and *unfamiliar*. In comparison, the *unfamiliar* images had different material constituents from the images involved in training, and they are used to test CNN's ability to segment new concrete scans in practice. Sample nomenclature indicates the sample name, type, and number (if applied), such as "Lime_train_1".

Table 1: Concrete samples investigated in this study.

| Mix | Name | Type | Total scan | Remarks on material constituent |
|-----|------|------|------------|--------------------------------|

| 1 | Lime | Train | 2 | Whitish limestone aggregate with a high sand ratio |
| --- | --- | --- | --- | --- |
| | | Test (*familiar*) | 1 | |
| 2 | Pebble | Train | 2 | High color variation in pebble aggregate |
| | | Test (*familiar*) | 1 | |
| 3 | Slag | Train | 2 | Grayish limestone aggregate with slag in paste |
| | | Test (*familiar*) | 1 | |
| 4 | Trap | Train | 2 | Traprock aggregate with silica fume in paste |
| | | Test (*familiar*) | 1 | |
| 5 | Flyash | Test (*unfamiliar*) | 1 | Larger limestone aggregate with fly ash in paste |
| 6 | Brown | Test (*unfamiliar*) | 1 | Brownish aggregate and paste |
| 7 | Cobble | Test (*unfamiliar*) | 1 | Cobble aggregate of various colors |
| 8 | Light | Test (*unfamiliar*) | 1 | Lightweight aggregate with a high void content |

The sample surface preparation generally followed a protocol previously reported by Song et al. unless otherwise specified [3]. After saw cutting each sample into a 60×60 mm flat specimen, the cross-section was progressively polished down to 1800 mesh (9 μm). The specimen was then air-dried in a desiccator until the visual effect brought by the surface moisture was removed. The air voids on the polishing section were then filled using an orange chalk powder (of controlled particle size around 1.6 μm), with the excess powder stricken off with a razor. Once these treatments were done, a digital image of the polished surface was collected using a flatbed scanner. Each of the scans was 50×50 mm, with a 5.3-μm pixel resolution.

Labels for the training images were obtained using an established colored-based method [3,28]. This required additional steps to process the training samples. First, the existing powder in the air voids was removed using compressive air. Then, a phenolphthalein solution (5 wt.% in 200 proof ethanol) was sprayed on the surface to dye the paste to pink. To minimize the potential negative influence from the color-based method, any sample showing major flaws were rejected and reprocessed from the polishing step. After drying, the air voids were filled again with the same chalk powder. Lastly, a second image for each training sample with the color treatment was scanned, where the scanning region was cautiously aligned to the first scan. In subsequent image processing, each pair of scans were further aligned in ImageJ to ensure a pixel-level agreement. The label images were segregated using MultiSpec. For aggregates and air voids, objects respectively smaller than 10000 and 100 μm$^2$ likely to be noise were ignored [3].

*3.2 CNN selection and model training*

This work adopted a CNN framework compiled by Huang et al. [45] that incorporates several latest advancements. The prototype of this framework is publicly available on GitHub [46]. Built on a state-of-art CNN DeepLabv3 [47] and with ResNet-101 [36] as the backbone, this framework has been tested in several studies with exceptional performance on prevailing benchmarks [48–51]. The algorithm implementation is conducted with PyTorch, an open-source machine learning library in Python.

The CNN model was trained for 20000 iterations to ensure sufficient optimization, and the training was accomplished with four graphics processing units (GPU). In each iteration, a batch of eight 800×800 pixel sections and their labels were randomly cropped from the training dataset (see Table 1). A jitter (i.e., a random combination of flipping, rotation, and scaling modification) was performed on the cropped sections to enrich the data variety. Thus, a total of 160000 sections were used for training. After each iteration, the CNN model parameters were updated using a stochastic gradient descent (SGD) optimizer. Additionally, a loss factor indicating the discrepancy between the model prediction and the label was updated to trace the model performance. The loss factor considers both the cross-entropy and Lovász-Softmax [52]. The entire work took about 72 hours, for finishing the 20000 training iterations.

*3.4 CNN model testing*

After training, the trained CNN model was used to process the testing images (see Table 1). Note that all these images were not processed using the color treatment. Timewise, segmenting a new concrete scan using the trained model was accomplished in a few seconds.

*3.4 Accuracy assessment*

The accuracy of the segmented image was statistically evaluated based on manual recognition. For each assessment, an orthogonal grid system was assigned to pinpoint 100×100 points across the entire concrete scan, and the same grid was replicated on the segmentation. To get the most reliable ground truth data of the sample, the 10000 points were manually annotated with the three phases—in a similar fashion to the manual point-counting method as specified in the ASTM C457. Then, the ground truth annotation was compared with the segmentation to obtain a confusion matrix from the 10000 points. Thus, the IoU of each the three phases and mIoU of the whole segmentation were calculated for indicating the segmentation accuracy.

Furthermore, the ASTM C457 parameters based on the different segmentations of each sample were computed using the point-counting method. This comparison was implemented to clarify the implication of the IoU accuracy to the air void parameters used in practical testing. This study focuses on the results of air content, paste content, and spacing factor.

**4. Results and discussion**

*4.1 Basic outputs of the CNN model*

The segmentation performance was traced with iterations using both the segmentation result and loss factor. Taking Lime_train_1 as an example, Fig. 4 displays its segmentations at three 50, 100, and 20000 iterations, with the aggregate shown in purple, paste in green, and air void in yellow. In Fig. 4b and c, the CNN model at the early iterations respectively learned identifying the air voids and further isolating a small portion of the paste. The quality of the ultimate segmentation is self-explanatory in Fig. 4d. Remarkably, the CNN model is not seemingly misled by the color variation of the aggregates.

The magnitude of loss correlates with the discrepancy between the training target (label) and training result (prediction), with 0 for a perfect match. It is often used as an indirect inference to the segmentation accuracy [42,44,49]. The loss recorded during training is displayed in Fig. 5, where a continuous decrease is seen. The rate of loss reduction becomes quite small after 1500 iterations, as the subsequent improvement mainly happened on refining the phase boundary. As a comparison, the segmentation accuracy of the CNN segmentations on the training dataset is additionally plotted in Fig. 5. A good reverse-correlation is observed between loss and accuracy. Note that the loss is calculated based on the discrepancy from the label images obtained from the color method, while the accuracy uses the ground truth data from human judgment as of the reference.

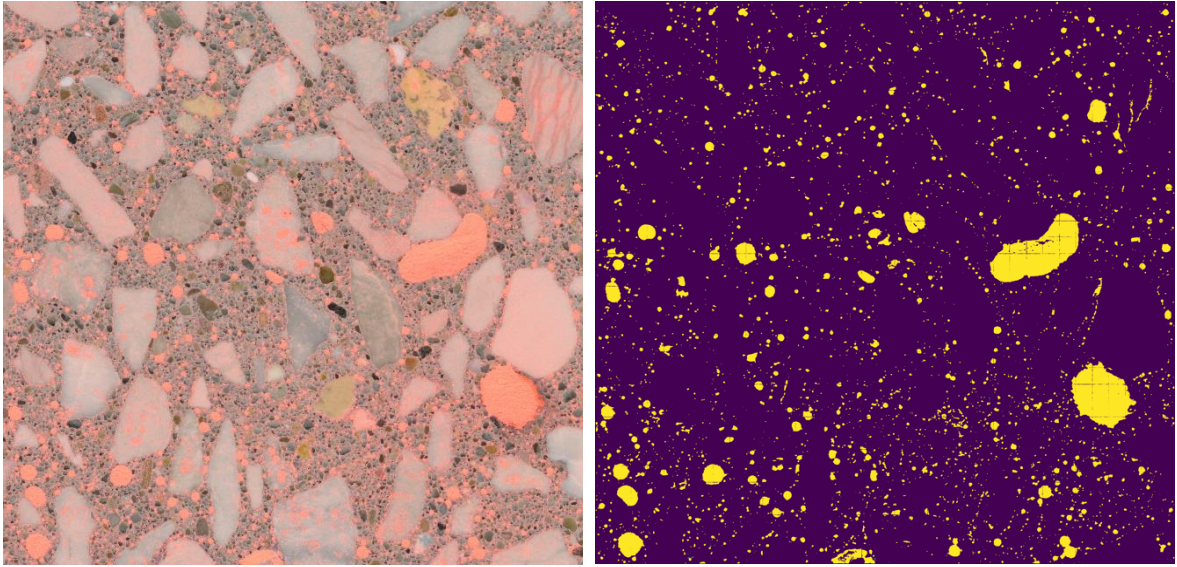

(a)            (b)

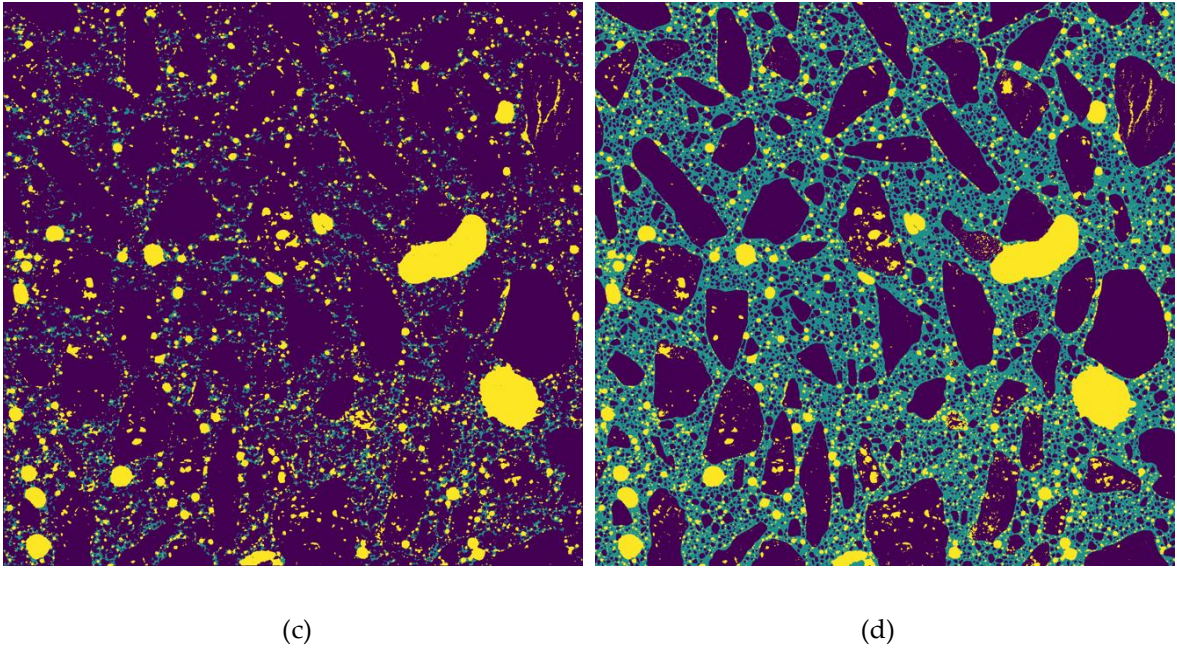

(c)                                          (d)

Fig. 4: CNN segmentation for Lime_train_1: (a) the 50×50 mm uncolored scan, and segmented images at (b) 50, (c) 100, and (d) 20000 iterations.

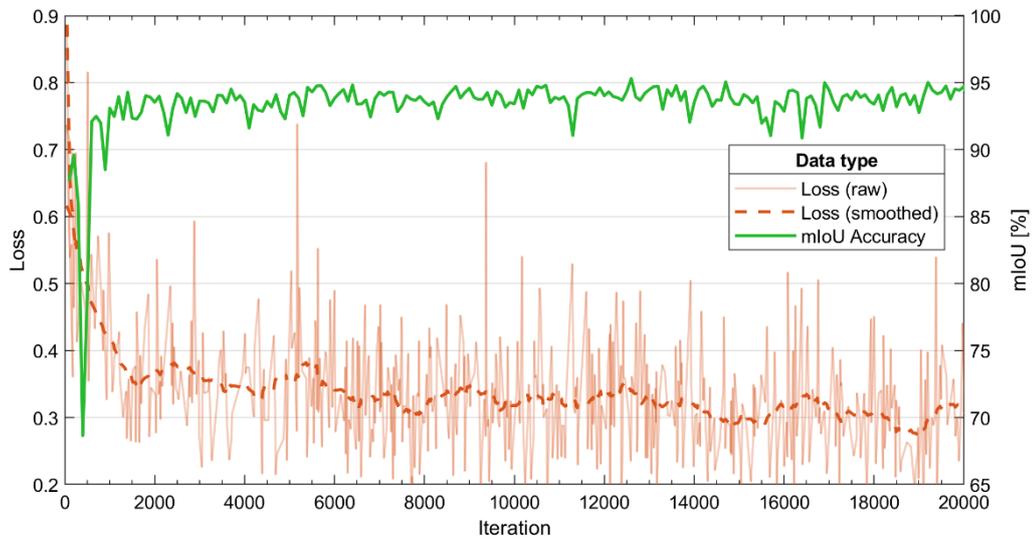

Fig. 5: The loss and accuracy of the CNN model training. This loss factor considered both the cross-entropy and Lovász-Softmax with equal weight. The accuracy here indicates the mIoU averaged from all training images.

*4.2 CNN vs. color-based segmentation for training dataset*

As the label images are obtained using the color method, a concern raised is that the CNN model can be affected by the flaws in the label images, so it is important to clarify its actual influence on CNN segmentation with accuracy assessment. Still taking "Lime_train_1" as an example, the 5×5 mm upper left corner of its label image and CNN segmentation at 20000 iterations are compared in Fig.

6. Fig. 6a shows the original scan, overlaid with cross markers highlighting the grid system used for the accuracy assessment. Fig. 6c displays the label image interpreted from Fig. 6b. In this color segmentation, some aggregates are mistakenly recognized as paste or bridged together. The high sand ratio of "Lime" concrete further leaves a greater challenge for the segmentation. However, these issues are not observed from the CNN segmentation in Fig. 6d, which is obtained directly from the uncolored scan. Despite the fact that some object boundary looks fuzzy in Fig. 6a, CNN yields a more precise description of the different phases. The above observations also hold for the other training images.

The accuracy statistics of Lime_train_1 are given in Table 2. The IoU accuracy results were calculated based on the confusion matrix of the 10000 grid points, as detailed in Section 3.4. A significant improvement can be seen in the CNN segmentation, where the mIoU raises from 0.872 to 0.929. As for the improvement in the individual phases, the greatest improvement happens to the differentiation between aggregate and paste. Interestingly, it seems that the CNN segmentation was able to capture the valid features of different phases. This may be explained by the fact that some flaws in the labels can be canceled out by the recurrence of similar yet correctly segmented features during training.

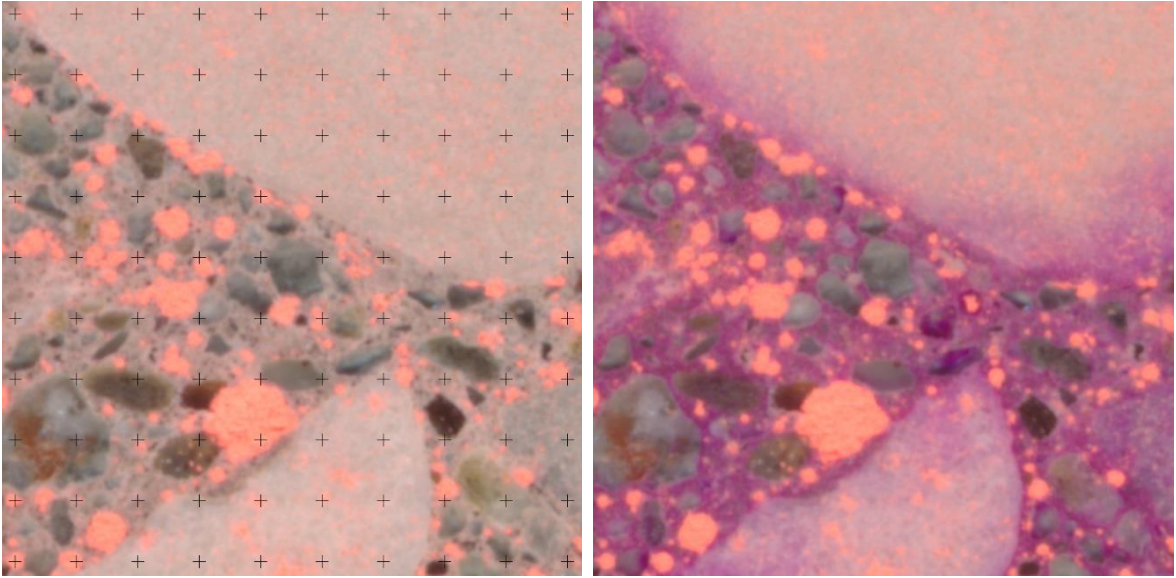

(a)          (b)

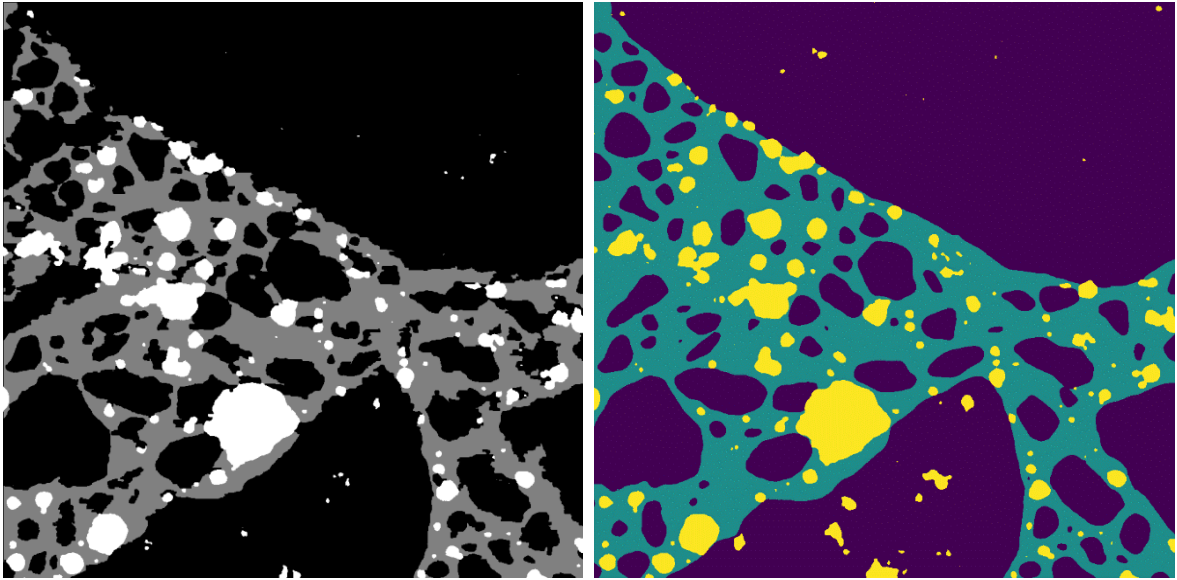

(c)             (d)

Fig. 6: A comparison on the 5×5 mm upper left corner of Lime_train_1: (a) the original sample surface, (b) same surface with the color treatment, (c) segmentation obtained with color-based analysis, and (d) CNN segmentation. The cross markers on (a) indicate the grid points used for the accuracy assessment.

Table 2: Confusion matrix and IoU accuracy of Lime_train_1.

| Confusion matrix | | Color-based | | | | | CNN | | | | |
|---|---|---|---|---|---|---|---|---|---|---|---|
| | | Segmentation | | | Accuracy | | Segmentation | | | Accuracy | |
| | | Agg. | Paste | Void | IoU | mIoU | Agg. | Paste | Void | IoU | mIoU |
| Ground truth | Agg. | 5697 | 227 | 5 | 0.900 | | 5673 | 251 | 23 | 0.948 | |
| | Paste | 369 | 2611 | 46 | 0.800 | 0.872 | 36 | 2964 | 15 | 0.901 | 0.929 |
| | Void | 34 | 8 | 1003 | 0.915 | | 3 | 254 | 1010 | 0.939 | |

The accuracy assessment result of each training image is summarized in Table 3, with the mIoU results further compared in Fig. 7. Due to the variations in material nature, the "Slag" and "Trap" samples show higher IOU accuracy than the other samples. As for the IoU for the individual phases, the accuracy of paste is lowest for most cases. The mIoU of the CNN segmentation is consistently higher than the color segmentation across the samples, where the averaged mIoU improved from 0.898 to 0.934. The higher accuracy of the CNN segmentation is also reflected in the ASTM C457 parameters inspected—air content $A$, paste content $P$, and spacing factor $L$. As compared with the IoU results, the C457 parameters show a smaller difference from the ground truth values. From a practical viewpoint, the C457 results from both the segmentation can provide a reasonable indication of the freeze-thaw performance. This can be attributed to that the IoU approach is more rigorous for highlighting the difference, while the C457 test is less affected by the precise phase detection on a pixel level.

Table 3: Segmentation accuracy for all training images.

| Sample | Type | IoU | | | mIoU | ASTM C457 | | |
|---|---|---|---|---|---|---|---|---|
| | | Agg. | Paste | Void | | A [%] | P [%] | L [um] |
| Lime_train_1 | Ground truth | 1.000 | 1.000 | 1.000 | 1.000 | 10.4 | 30.1 | 0.136 |
| | color-based | 0.900 | 0.800 | 0.915 | 0.872 | 10.5 | 28.5 | 0.127 |
| | CNN | 0.948 | 0.901 | 0.939 | 0.929 | 10.5 | 32.4 | 0.146 |
| Lime_train_2 | Ground truth | 1.000 | 1.000 | 1.000 | 1.000 | 11.4 | 32.1 | 0.119 |
| | color-based | 0.869 | 0.740 | 0.887 | 0.832 | 10.6 | 27.9 | 0.124 |
| | CNN | 0.924 | 0.884 | 0.916 | 0.908 | 11.8 | 35.4 | 0.131 |
| Pebble_train_1 | Ground truth | 1.000 | 1.000 | 1.000 | 1.000 | 10.5 | 29.8 | 0.089 |
| | color-based | 0.940 | 0.865 | 0.812 | 0.872 | 10.2 | 31.1 | 0.092 |
| | CNN | 0.942 | 0.882 | 0.904 | 0.909 | 10.2 | 30.9 | 0.093 |
| Pebble_train_2 | Ground truth | 1.000 | 1.000 | 1.000 | 1.000 | 6.7 | 33.1 | 0.266 |
| | color-based | 0.937 | 0.910 | 0.936 | 0.927 | 5.1 | 38.2 | 0.203 |
| | CNN | 0.946 | 0.911 | 0.933 | 0.930 | 6.6 | 35.0 | 0.269 |
| Slag_train_1 | Ground truth | 1.000 | 1.000 | 1.000 | 1.000 | 11.1 | 29.1 | 0.095 |
| | color-based | 0.934 | 0.852 | 0.936 | 0.907 | 11.1 | 26.3 | 0.085 |
| | CNN | 0.954 | 0.901 | 0.958 | 0.938 | 11.1 | 27.3 | 0.089 |
| Slag_train_2 | Ground truth | 1.000 | 1.000 | 1.000 | 1.000 | 8.0 | 29.5 | 0.150 |
| | color-based | 0.932 | 0.860 | 0.937 | 0.910 | 8.1 | 32.2 | 0.164 |
| | CNN | 0.971 | 0.938 | 0.970 | 0.960 | 7.9 | 30.3 | 0.154 |
| Trap_train_1 | Ground truth | 1.000 | 1.000 | 1.000 | 1.000 | 9.5 | 33.7 | 0.220 |
| | color-based | 0.947 | 0.903 | 0.932 | 0.928 | 8.9 | 34.2 | 0.246 |
| | CNN | 0.962 | 0.929 | 0.947 | 0.946 | 9.7 | 34.1 | 0.222 |
| Trap_train_2 | Ground truth | 1.000 | 1.000 | 1.000 | 1.000 | 8.2 | 37.0 | 0.092 |
| | color-based | 0.933 | 0.907 | 0.973 | 0.938 | 8.2 | 39.0 | 0.094 |
| | CNN | 0.965 | 0.952 | 0.935 | 0.951 | 8.0 | 38.3 | 0.092 |
| Average | color-based | 0.924 | 0.855 | 0.916 | 0.898 | - | - | - |
| | CNN | 0.952 | 0.912 | 0.938 | 0.934 | - | - | - |

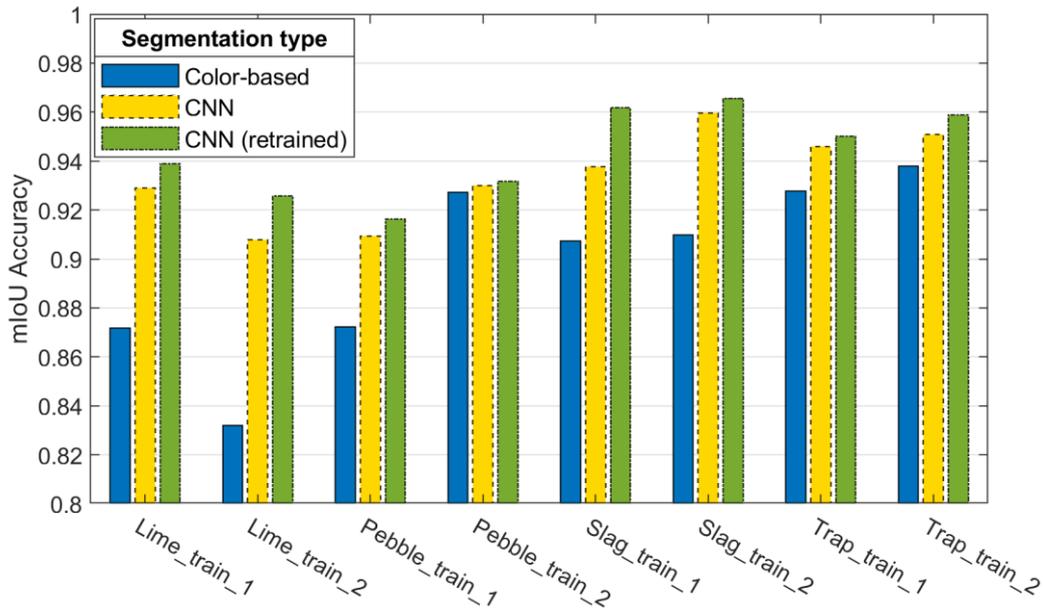

Fig. 7: A comparison on mIoU accuracy for the samples in the training dataset with different types of segmentation. [Note, the results of "CNN (retain)" are detailed in Section 4.3.]

*4.3 Retraining for accuracy improvement*

From the training results shown above, it is noticed that a few labels have lower accuracy, which may affect the training quality. Given the higher accuracy of the segmented images by CNN, a possible way to fix this issue is replacing the low-quality labels with the segmented images by CNN and run another round of model training. This idea was implemented on Lime_train_1, Lime_train_2, Pebble_train_1, and Slag_train_1. The rationale for keeping the other labels is allowing CNN to learn critical features from the color-based segmentation, which may not be fully reflected from the CNN segmentation that has been obtained. The new accuracy results obtained from the retraining of 20000 iterations are given in Table 4. In comparison (see Fig. 7), a moderate further improvement is achieved by the new model, with the averaged mIoU increased by 0.01 from the original CNN model.

Table 4: Segmentation accuracy for all training images after retraining.

| Sample | Type | IoU | | | mIoU | ASTM C457 | | |
|---|---|---|---|---|---|---|---|---|
| | | Agg. | Paste | Void | | A [%] | P [%] | L [um] |
| Lime_train_1 | CNN (retrain) | 0.958 | 0.916 | 0.943 | 0.939 | 10.3 | 31.9 | 0.143 |
| Lime_train_2 | CNN (retrain) | 0.941 | 0.903 | 0.933 | 0.926 | 11.5 | 34.6 | 0.128 |
| Pebble_train_1 | CNN (retrain) | 0.947 | 0.891 | 0.911 | 0.916 | 10.2 | 30.9 | 0.093 |
| Pebble_train_2 | CNN (retrain) | 0.951 | 0.920 | 0.924 | 0.932 | 6.4 | 34.6 | 0.265 |
| Slag_train_1 | CNN (retrain) | 0.976 | 0.947 | 0.962 | 0.962 | 11.0 | 29.7 | 0.097 |
| Slag_train_2 | CNN (retrain) | 0.977 | 0.949 | 0.971 | 0.966 | 7.9 | 29.9 | 0.152 |
| Trap_train_1 | CNN (retrain) | 0.966 | 0.934 | 0.951 | 0.950 | 9.6 | 34.0 | 0.222 |
| Trap_train_2 | CNN (retrain) | 0.971 | 0.960 | 0.945 | 0.959 | 7.9 | 37.7 | 0.091 |
| Average | CNN (retrain) | 0.961 | 0.928 | 0.942 | 0.944 | - | - | - |

*4.4 CNN segmentation for testing dataset*

Different from the training images discussed above, the results on the testing images reflect a more realistic segmentation performance, as these images are not involved in training. For the two subgroups of the testing images investigated (see Table 1), the *familiar* portion contains the same concrete types as for training images, while the *unfamiliar* portion only has new concrete types, which are more challenging to analyze. Fig. 8 displays the testing images segmented by the retrained CNN model at 20000 iterations. For each segmentation, the inset section shows a 5×5 mm local magnification of its upper left corner. Macroscopically, the CNN segmentation gives a proper phase description for the original testing scans, including the *unfamiliar* samples. The appearance of aggregates and cement paste varies widely across the samples and even within the same scan, whereas the CNN segmentation is not much affected.

The CNN segmentation works even with challenging samples. Even semi-transparent quartz sand and blurry phase transitions are well isolated in the CNN segmentation. As seen from Slag_test in Fig. 8c, Brown_test in Fig. 8f, and Cobble_test in Fig. 8g, CNN can reasonably separate phases even with almost the same color. Note Light_test in Fig. 8g contains lightweight aggregates. Based on our previous experience, conducting color-based segmentation is infeasible for this kind of sample, due to its high absorption of color dye during surface treatment [3]. However, this issue can be avoided by handling the uncolored samples directly using CNN.

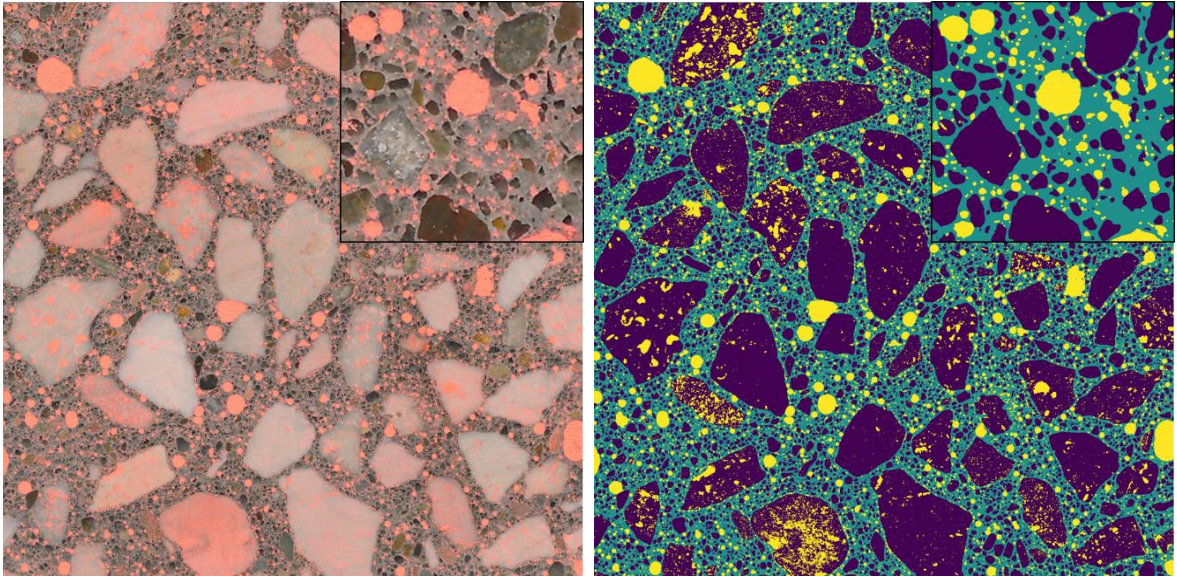

(a)

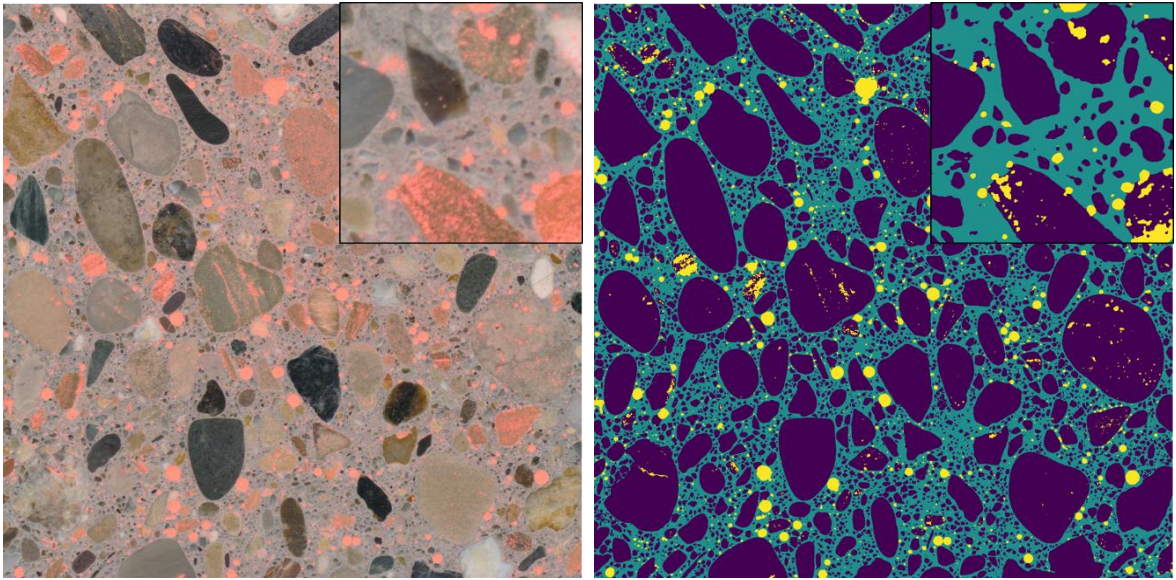

(b)

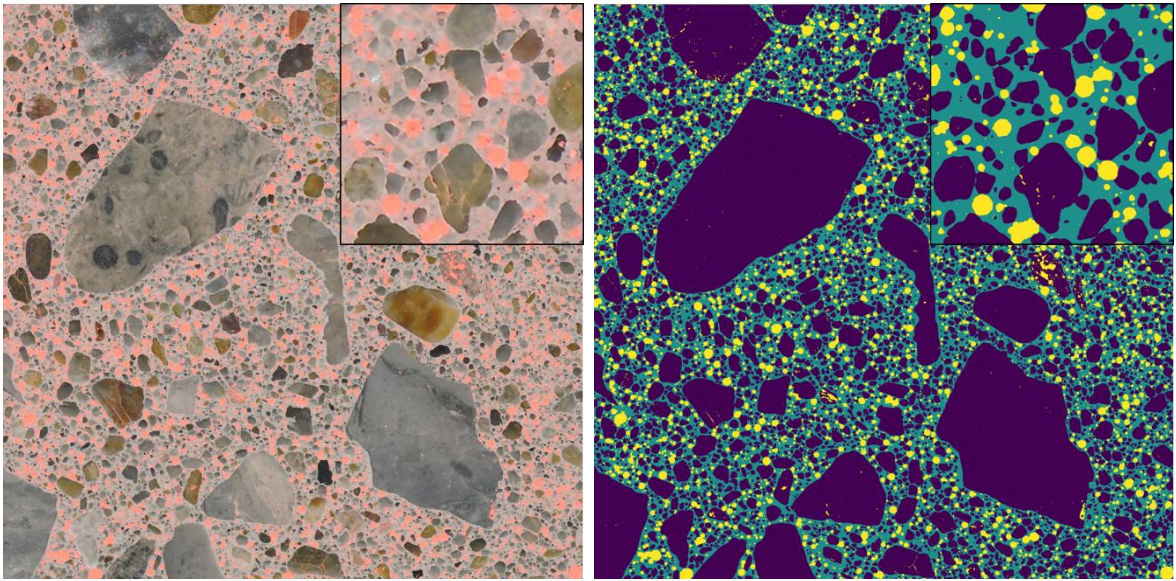

(c)

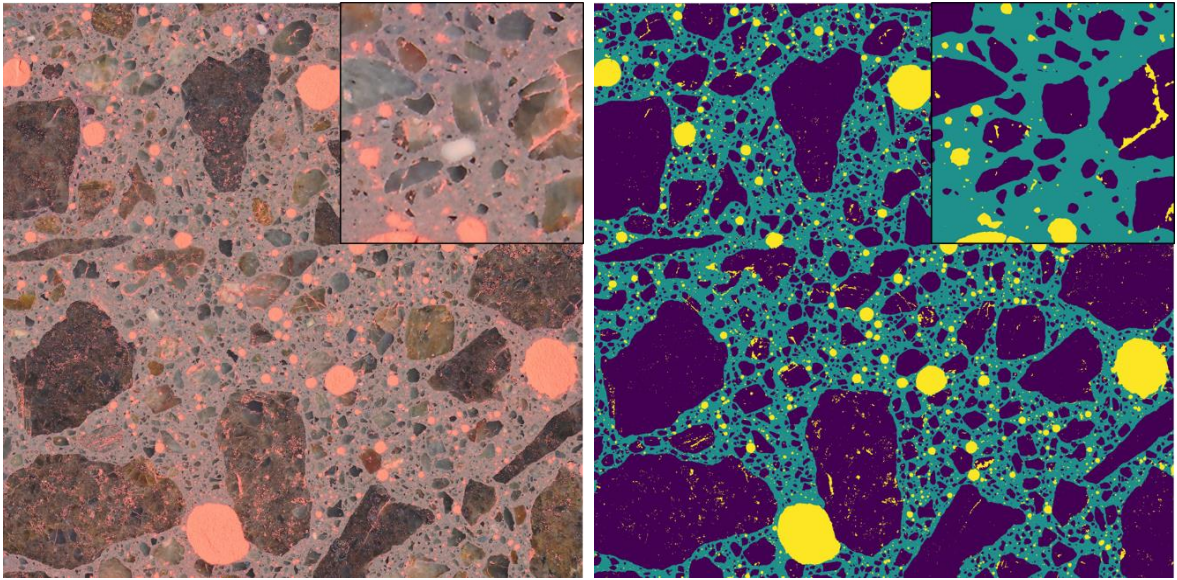

(d)

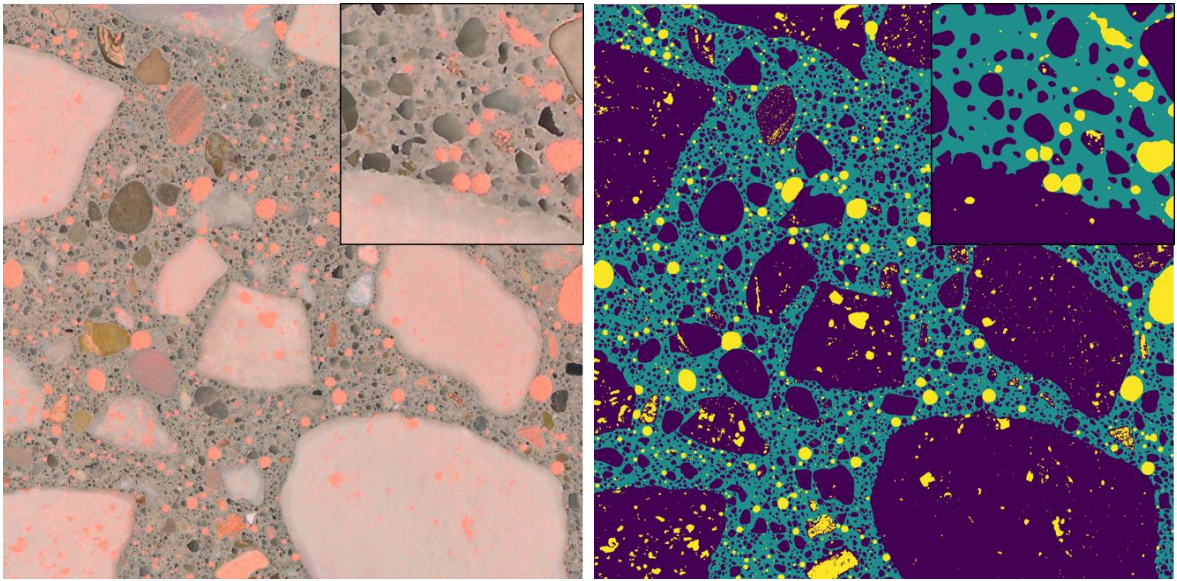

(e)

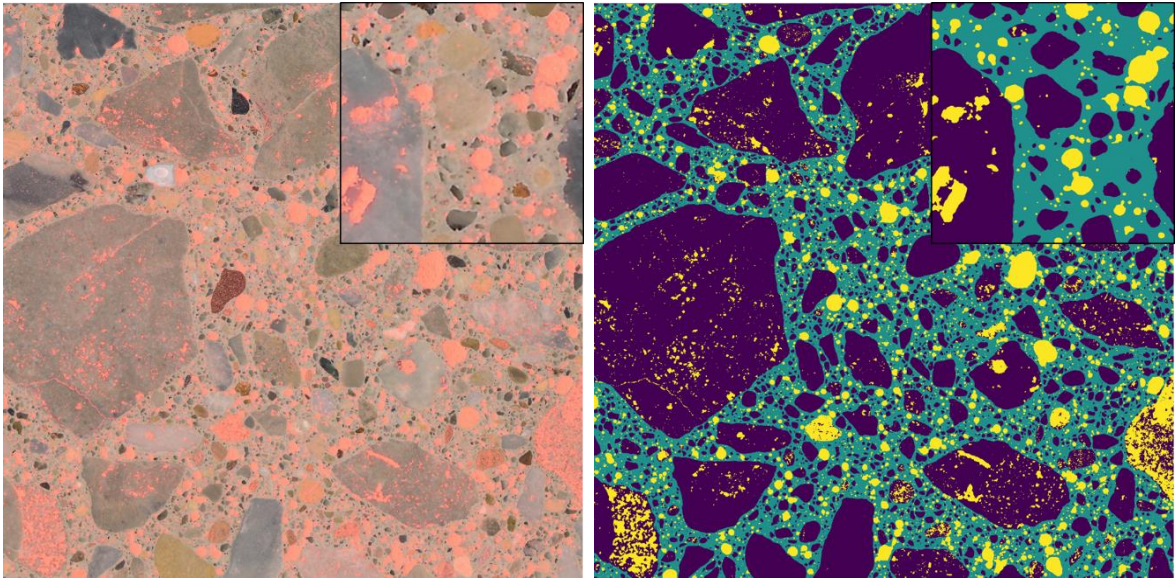

(f)

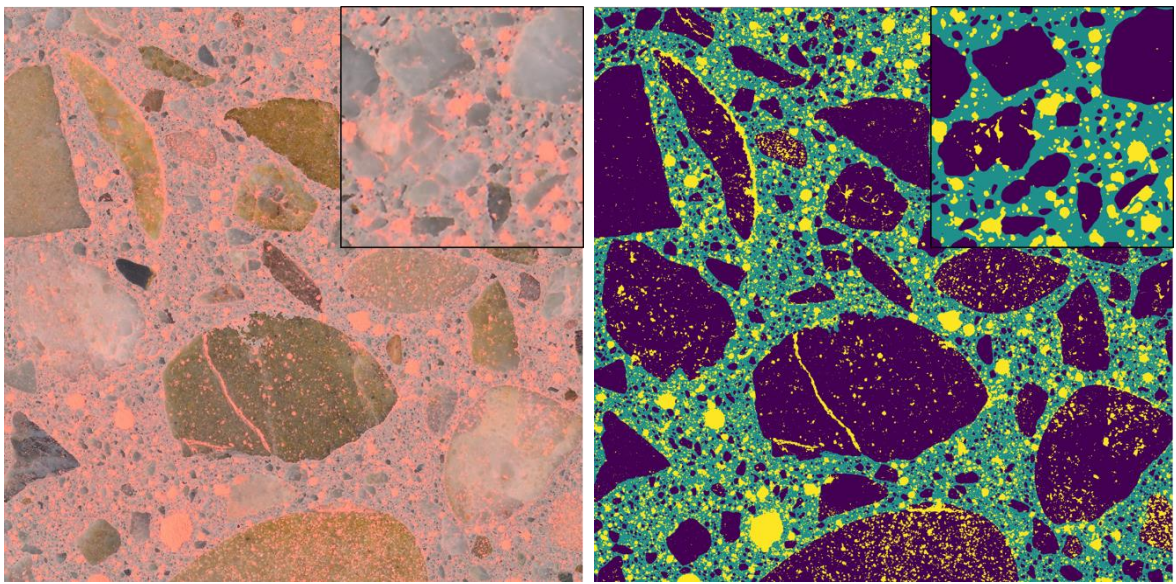

(g)

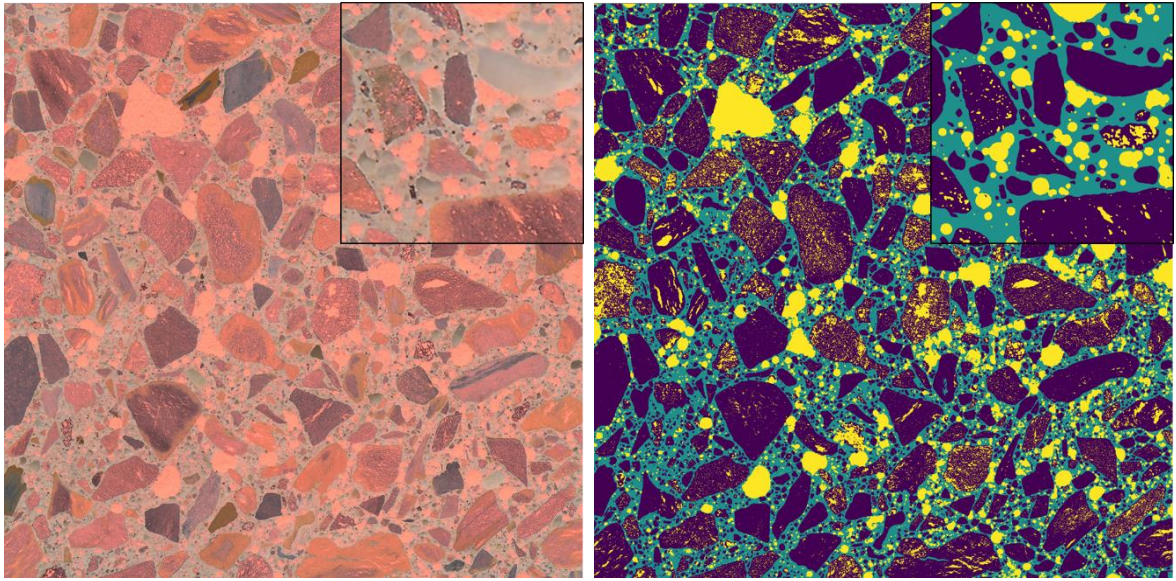

(h)

Fig. 8: CNN segmentation of the testing images: (a) Lime_test, (b) Pebble_test, (c) Slag_test, (d) Trap_test, (e) Flyash_test, (f) Brown_test, (g) Cobble_test, (h) Light_test. Each of the inset sections magnifies the 5×5 mm upper left corner of the original 50×50 mm scan.

The accuracy statistics for the testing images are summarized in Table 5. Compared with the training images (Table 4), the averaged mIoU of the testing images is slightly diminished from 0.944 to 0.934. Due to the higher confidence level for CNN to segment *familiar* testing images, the averaged mIoU of the *familiar* samples is higher than that of the *unfamiliar* samples by 0.011. Interestingly, Flyash_test (Fig. 8e) that has the lowest mIoU, 0.894, does not exhibit the worst visual contrast. In its segmentation, it can be found that some coarse aggregates are merged with the nearby fine aggregates. This should be a representativeness issue of the CNN model, which can be fixed by further enriching concrete types of training dataset. Consistent with the training images, the C457 results given by the CNN segmentation for the testing images are close to the ground truth values. Even for Flyash_test, its C457 results measured based on the CNN segmentation are rather reasonable.

Table 5: Segmentation accuracy for the testing images.

| Sample | Type | IoU | | | mIoU | ASTM C457 results | | |
|---|---|---|---|---|---|---|---|---|
| | | Agg. | Paste | Void | | A [%] | P [%] | L [um] |
| Lime_test (familiar) | Ground truth | 1.000 | 1.000 | 1.000 | 1.000 | 13.3 | 29.0 | 0.053 |
| | CNN | 0.932 | 0.883 | 0.926 | 0.913 | 13.0 | 30.7 | 0.056 |
| Pebble_test (familiar) | Ground truth | 1.000 | 1.000 | 1.000 | 1.000 | 3.8 | 34.8 | 0.250 |
| | CNN | 0.967 | 0.941 | 0.922 | 0.943 | 4.0 | 34.8 | 0.255 |
| Slag_test | Ground truth | 1.000 | 1.000 | 1.000 | 1.000 | 9.2 | 32.3 | 0.149 |
| (familiar) | CNN | 0.967 | 0.930 | 0.932 | 0.943 | 9.0 | 33.1 | 0.153 |
| Trap_test (familiar) | Ground truth | 1.000 | 1.000 | 1.000 | 1.000 | 7.9 | 35.9 | 0.126 |
| | CNN | 0.970 | 0.953 | 0.950 | 0.958 | 7.7 | 36.4 | 0.125 |
| Flyash_test (unfamiliar) | Ground truth | 1.000 | 1.000 | 1.000 | 1.000 | 8.6 | 32.6 | 0.146 |
| | CNN | 0.911 | 0.867 | 0.902 | 0.894 | 7.9 | 35.2 | 0.155 |
| Brown_test (unfamiliar) | Ground truth | 1.000 | 1.000 | 1.000 | 1.000 | 13.8 | 28.1 | 0.100 |
| | CNN | 0.966 | 0.929 | 0.952 | 0.949 | 14.0 | 28.3 | 0.100 |
| Cobble_test (unfamiliar) | Ground truth | 1.000 | 1.000 | 1.000 | 1.000 | 16.8 | 28.7 | 0.045 |
| | CNN | 0.958 | 0.921 | 0.940 | 0.939 | 16.3 | 29.9 | 0.047 |
| Light_test (unfamiliar) | Ground truth | 1.000 | 1.000 | 1.000 | 1.000 | 17.7 | 26.2 | 0.045 |
| | CNN | 0.945 | 0.921 | 0.921 | 0.929 | 17.4 | 27.4 | 0.047 |
| Average | Familiar | 0.959 | 0.927 | 0.932 | 0.939 | - | - | - |
| | Unfamiliar | 0.945 | 0.909 | 0.929 | 0.928 | - | - | - |
| | Overall | 0.952 | 0.918 | 0.931 | 0.934 | - | - | - |

*4.5 Potential of deep learning in broader petrographic applications*

This study demonstrates that machine visual understanding can be a powerful tool for concrete petrographic analysis. The use of CNN makes it possible to segment concrete scans without color enhancement. CNN has been shown to distinguish the different phases in the uncolored concrete scan. In many cases shown in this paper, the CNN segmentation approaches the quality achieved by human judgment, but at a small fraction of the required time. CNN segmentations are complete in seconds, and with a simple program script, ASTM C457 parameters can be computed immediately.

Broadly, the accurate visual understanding of the material composition opens new possibilities for predicting various concrete performance conveniently and expeditiously. By applying deep learning techniques to petrography, the benefits may not only be higher accuracy and less time, but also understanding the material property and behavior from a novel perspective. In future investigations, it would also be interesting to explore the potential use of deep learning for other concrete applications, such as structural health monitoring or non-destructive testing.

**5. Conclusion**

This study explored the potential of using novel deep learning techniques for concrete petrographic analysis. The CNN method was shown to segment concrete images without the use of phenolphthalein to add color to cement paste. Two groups of concrete scans were prepared, with one for CNN model training and the other for performance testing. For each scan, a rigorous manual recognition was conducted to obtain a reliable reference for evaluating the segmentation accuracy based on the IoU index and C457 parameters. Compared to the color-based segmentation, the CNN segmentation achieved considerably higher accuracy on the training images. The CNN method also exhibited unprecedented performance on the testing images, which include several new concrete types never involved in the model training. For some samples, CNN was even competitive against human judgment. As compared with the IoU results, the C457 parameters calculated from the CNN segmentation were closer to the ground truth results.

The results demonstrate that CNN has strong accuracy and time advantages over the conventional color-based approach. Taking the ASTM C457 analysis as an example, the total processing time to output the final air void parameters from a concrete scan is successfully reduced from hours of manual inspection to mere seconds. In short, CNN can be a powerful tool for concrete petrographic analysis, and the accurate machine visual understanding further opens up many new possibilities for concrete research.

**Acknowledgment**

This work utilizes resources supported by the National Science Foundation's Major Research Instrumentation program, grant #1725729, as well as the National Center for Supercomputing Applications at University of Illinois Urbana-Champaign. The authors thank Anne Werner for valuable discussion and suggestions. The authors would also thank Mengxiao Zhong, Chang Chen, Yilin Ma, and Xuhuan Zhao for their time and effort involved with the manual image analysis.